\documentclass[10pt,twocolumn,letterpaper]{article}

\usepackage{iccv}
\usepackage{times}
\usepackage{epsfig}
\usepackage{graphicx}
\usepackage{amsmath}
\usepackage{amssymb}

\usepackage{booktabs}
\usepackage{float}
\usepackage{subfig}
\usepackage{multirow}
\usepackage{multicol}
\usepackage{enumitem}

\usepackage{amssymb}
\usepackage{amsmath}
\usepackage{pifont}
\usepackage{graphbox} 
\usepackage{xcolor,soul}

\usepackage[pagebackref=true,breaklinks=true,colorlinks,bookmarks=false]{hyperref}

\iccvfinalcopy 

\ificcvfinal\fi
\begin{document}

\title{SportsSloMo: A New Benchmark and Baselines for \\Human-centric Video Frame Interpolation}

\author{Jiaben Chen\thanks{Work mainly done when Jiaben Chen was an intern at Northeastern University.}\\
UC San Diego\\
{\tt\small jic088@ucsd.edu}
\and
Huaizu Jiang\\
Northeastern University\\
{\tt\small h.jiang@northeastern.edu}
}
\maketitle

\ificcvfinal\thispagestyle{empty}\fi

\begin{abstract}
Human-centric video frame interpolation has great potential for enhancing entertainment experiences and finding commercial applications in the sports analysis industry, \eg, synthesizing slow-motion videos.
Although there are multiple benchmark datasets available for video frame interpolation in the community, none of them is dedicated to human-centric scenarios.
To bridge this gap, we introduce \textbf{SportsSloMo}, a benchmark featuring over 130K high-resolution ($\geq$720p) slow-motion sports video clips, totaling over 1M video frames, sourced from YouTube.
We re-train several state-of-the-art methods on our benchmark, and we observed a noticeable decrease in their accuracy compared to other datasets. This highlights the difficulty of our benchmark and suggests that it poses significant challenges even for the best-performing methods, as human bodies are highly deformable and occlusions are frequent in sports videos.
To tackle these challenges, we propose human-aware loss terms, where we add auxiliary supervision for human segmentation in panoptic settings and keypoints detection.
These loss terms are model-agnostic and can be easily plugged into any video frame interpolation approach.
Experimental results validate the effectiveness of our proposed human-aware loss terms, leading to consistent performance improvement over existing models.
The dataset and code can be found at: \href{https://neu-vi.github.io/SportsSlomo/}{https://neu-vi.github.io/SportsSlomo/}.
\end{abstract}

\section{Introduction \label{1}}
Video frame interpolation (VFI) is a technique that synthesizes intermediate frames from input images, enhancing the clarity of content that may be difficult to see otherwise. 
This technique finds wide-ranging applications, including slow-motion video generation~\cite{jiang2018superslomo}, novel view synthesis~\cite{li2021neural}, video compression~\cite{wu2018video}, cartoon and rendered content generation~\cite{li2021deep,briedis2021neural}, etc.
In recent years, we have witnessed significant advances in this field, due in large part to the development of various benchmarks~\cite{baker2011database,soomro2012ucf101,perazzi2016benchmark,su2017deep,xue2019video,choi2020channel,sim2021xvfi}.

Humans feature prominently in most contemporary videos. With the widespread use of mobile devices, people can easily record and share their daily experiences with family, friends, and colleagues.
Meanwhile, live broadcasts of sporting events attract a large audience. 
Automatically generated slow-motion videos can create a more immersive and engaging experience for users by highlighting details and valuable moments of their lives that may be missed in real-time. 
Therefore, improving video frame interpolation results for human-centric videos has great potential for enhancing user experiences in entertainment.
Human-centric video frame interpolation approaches can also be beneficial in various industries. 
Athletes and coaches, for example, can use slow-motion synthesis to identify flaws in techniques, highlight areas for improvement, and gain a more detailed understanding of how different factors may contribute to the success.
\begin{figure}[t]
    \centering
    \includegraphics[width=\linewidth]{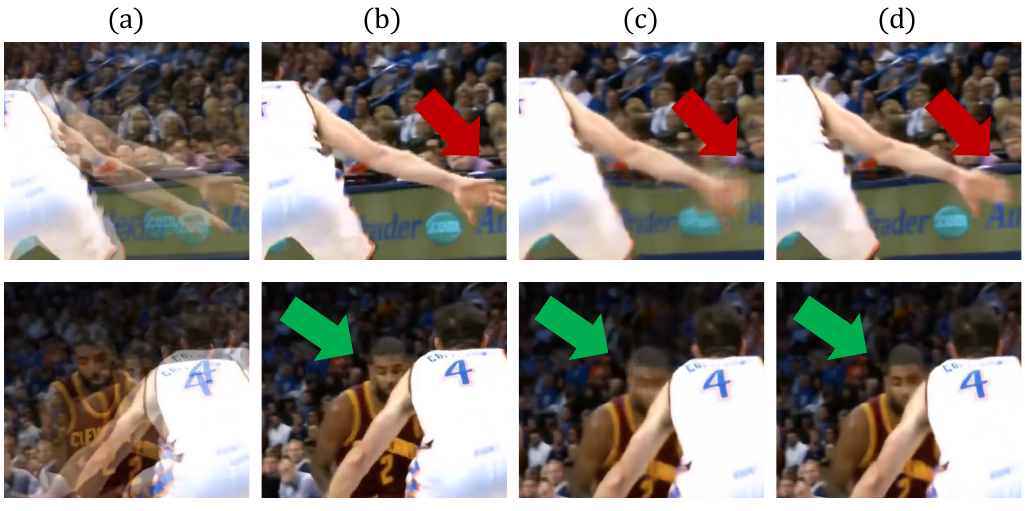}
    \caption{\textbf{Human-centric video frame interpolation results.} We propose human-aware auxiliary losses to improve interpolation accuracy at motion boundaries. From left to right: (a): Overlayed inputs, (b): Ground truth, (c): Interpolation results, (d): Interpolation results with our proposed human-aware loss terms.} 
    \vspace{-6pt}
    \label{fig:teaser}
\end{figure}

\textbf{A new benchmark.} Despite the availability of various benchmarks for video frame interpolation, a notable gap exists in datasets specifically tailored for human-centric scenarios.
To bridge this gap and to foster the research in this important direction, we introduce \textbf{SportsSloMo}, a new dataset comprising high-resolution ($\geq$720p) slow-motion sports videos crawled from YouTube under the Common Creative Licence. This dataset encompasses a diverse range of sports, such as football, basketball, baseball, hockey, etc.
Since a video may contain advertisement, transition frames, changes of shot, and non-slow-motion content, we carefully curate the data to remove such unwanted content and finally split each long video into a set of short slow-motion clips of 9 frames.
The first and last frames are used as input and the rest 7 intermediate frames are reserved as ground truths for training and evaluating VFI models.
In total, our benchmark has 130K video clips and more than 1M video frames. 
Compared with existing datasets, as shown in Table~\ref{tab:dataset_statistics}, our proposed SportsSloMo benchmark is the largest one so far, with high resolution and focus on human-centric scenarios.

While primarily designed for human-centric VFI in this paper, we believe SportsSloMo dataset may also have the potential to aid research in other tasks such as video super-resolution~\cite{chan2022investigating,yli_comisr_iccv2021}, group activity recognition~\cite{tamura2022hunting,zhou2022composer,yuan2021DIN}, and dynamic view synthesis~\cite{li2021neural,Gao2021DynNeRF,xian2021space}. By releasing this dataset to the entire community, we hope to encourage technical advancement in human-centric video frame interpolation and empower researchers to explore innovative applications in other adjacent fields.

\begin{figure}[t]
\centering
    \includegraphics[width=0.95\linewidth]{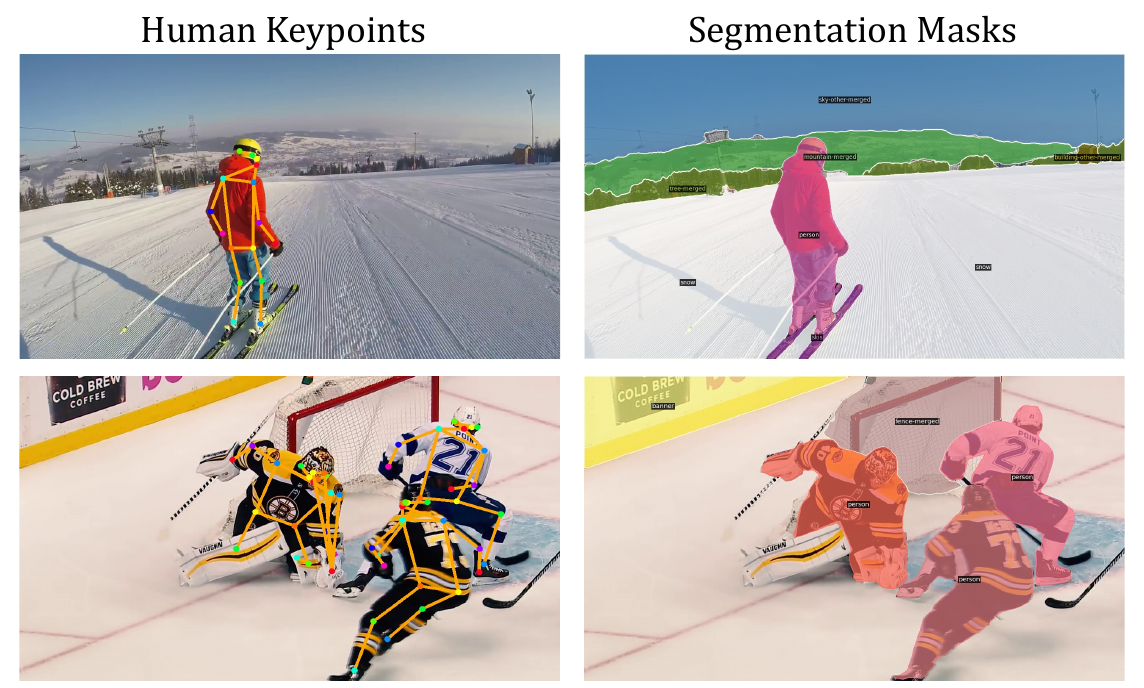}
\vspace{-6pt}
\caption{\textbf{Visualization of human keypoints~\cite{xu2022vitpose} and panoptic segmentation masks~\cite{cheng2022masked}.}}
\vspace{-10pt}
\label{fig:kptseg}
\end{figure}

\textbf{Benchmarking existing approaches.} To facilitate development and evaluation of human-centric video frame interpolation methods, we re-train several state-of-the-art approaches~\cite{jiang2018superslomo,lee2020adacof,hu2022manytomany,huang2022real,jin2023enhanced,jin2023unified,zhang2023extracting} using their publicly released code on our SportsSloMo dataset.
As the human bodies are highly deformable and occlusions are frequent in sports videos, the accuracy of all the methods decrease compared to their performance on other datasets.
For instance, EBME~\cite{jin2023enhanced} and EMA-VFI~\cite{zhang2023extracting}, two of the top-performing approaches on SportsSloMo, produce PSNR scores of 30.15 and 30.70, respectively - markedly lower than their scores of 36.19 and 36.64 on Viemo90K~\cite{xue2019video}, as well as 30.64 and 30.94 on the hard split of the SNU-FILM~\cite{choi2020channel} benchmarks.
It highlights the difficulty of our benchmark and suggests significant challenges need to be addressed.

\textbf{Enhancing models for human-centric VFI.} To improve the existing VFI models on our benchmark, we introduce human-aware priors to enhance the model training. 
Specifically, we propose loss terms based on human segmentation in the panoptic setting~\cite{cheng2022masked} and human keypoints estimation~\cite{xu2022vitpose} as extra supervision for intermediate frame synthesis. Fig. \ref{fig:kptseg} shows a visualization of detected human keypoints and segmentation masks in our dataset.
Human segmentation masks help delineate human body boundaries, which are helpful for reducing ghost effects around the motion boundaries. 
At the same time, human keypoints estimation can also indicate where each body part is, enforcing coherent motion trajectories in the synthesized video frames. 
Specifically, we compare the output from pre-trained panoptic segmentation and keypoints detection models taking input as a synthesized and the ground-truth intermediate frame, respectively, and use the consistency as supervision.
As shown in Fig. \ref{fig:teaser}, by supervising with our proposed human-aware loss terms, we improve the interpolation quality at motion boundaries with less blurry results in scenarios with large motion and occlusion.
Both of these human-aware loss terms are model agnostic and can be easily integrated into any video frame interpolation approach.
Experimental results show that they can consistently improve the accuracy of seven existing approaches, leading to strong baselines on our benchmark.

\newcommand{\cmark}{\ding{51}}%
\newcommand{\xmark}{\ding{55}}%
\begin{table}[t]
    \centering
    \caption{\textbf{Comparisons of different benchmark datasets for video frame interpolation.} (\#inter. frames indicates the number of intermediate frames to synthesize.)}
    \label{tab:dataset_statistics}
    \renewcommand{\tabcolsep}{1.5pt}
    \small
    \begin{tabular}{c|c|c|c|c|c}
    \toprule
    \multirow{2}{*}{Dataset} & \multirow{2}{*}{\#clips} & \multirow{2}{*}{\#images} & \#inter. & \multirow{2}{*}{resolution} & human- \\
     & & & frames & & centric \\
    \midrule
     UCF101 \cite{soomro2012ucf101} & 0.4K & 1K &1 & 256$\times$256 & \cmark\\
     Adobe240fps \cite{su2017deep} & 0.1K   &  80K & 7 & 1280$\times720$ & \scalebox{0.85}[1]{\xmark}\\
     Vimeo90K \cite{xue2019video} & 70K   &  220K & 1 & 448$\times$256 & \scalebox{0.85}[1]{\xmark} \\
     SNU-FILM \cite{choi2020channel} & 1.2K & 3.6K & 1 & 1280$\times720$ & \scalebox{0.85}[1]{\xmark}\\
     X4K1000FPS \cite{sim2021xvfi} & 4.4K & 286K & 7 & 4096$\times$2160 & \scalebox{0.85}[1]{\xmark}\\
    \midrule
    SportsSloMo & 130K & 1183K & 7 & 1280$\times$720 & \cmark\\
    \bottomrule
    \end{tabular}
\end{table}

To sum up, this paper makes the following contributions:
\setlength{\leftmargini}{0.85em}
\begin{itemize}[topsep=0pt,itemsep=0.75pt]
    \item We introduce SportsSloMo, a new benchmark dataset consisting of a large amount of slow-motion sports videos. To the best of our knowledge, this is the first high-resolution dataset tailored for human-centric video frame interpolation, supporting synthesis of multiple intermediate frames.
    \item We benchmark state-of-the-art approaches on the new benchmark, 
    highlighting the challenges of the human-centric video frame interpolation task.
    \item We propose two human-aware loss terms, which take the human priors into account for video frame interpolation and can easily be plugged into existing video frame interpolation approaches. Experimental results validate that they can consistently improve existing models, yielding strong baseline models on our new benchmark.
\end{itemize}

\section{Related Work}

\subsection{Benchmark datasets} 
Existing publicly available datasets already provide a valuable resource for developing and evaluating video frame interpolation methods. We would briefly introduce these datasets and reveal some limitations in this section. Table. \ref{tab:dataset_statistics} shows a comparison between existing datasets and our proposed SportsSloMo dataset.

The SNU-FILM \cite{choi2020channel} dataset is a widely-used benchmark for VFI evaluation, containing 1240 frame triplets of 1280 $\times$ 720 resolution. And it is divided into four different parts, namely, Easy, Medium, Hard, and Extreme according to motion magnitude. The Middlebury benchmark \cite{baker2011database} is another widely used dataset, image resolution in this dataset is around $640 \times 480$. However, it is commonly only used to evaluate VFI methods for 8 sequences. UCF101 \cite{soomro2012ucf101} is originally a dataset for human action recognition, containing a variety of human actions. With the test set constructed by \cite{liu2017video}, it is also used to evaluate VFI methods, containing 379 triplets of 256 $\times$ 256 frame size. Nevertheless, its scale is rather small and its resolution is also low. The Adobe240fps dataset \cite{su2017deep}, originally for video deblurring, is another widely used dataset for VFI. It is consisted of high frame-rate videos (240 fps) with a resolution of 1280 $\times$ 720, yet the videos are from only 118 clips. 

The mostly used dataset for training and evaluating VFI methods is the Vimeo90K dataset \cite{xue2019video}. It contains 73171 frame triplets from 14777 video clips extracted from real-life video clips with a fps $\le 30$. Nevertheless, one of its main drawback is the low resolution of 448 $\times$ 256 obtained by downscaling the original high resolution frames. Moreover, as VFI methods have been rapidly improving in recent years, their performance on the widely-used Vimeo90K dataset has approached saturation. To further advance the state-of-the-art in video frame interpolation, a new dataset with bigger scale, higher resolution and more challenging scenarios is necessary. X4K1000FPS is a recently released high frame rate (1000 fps) with 4K spatial resolution to promote the study of VFI for very high resolution videos.

As a result, none of existing datasets contains rich human-centric data with high resolution and big scale. To bridge this gap, we introduce the SportsSloMo dataset, a human-centric VFI dataset with 130K video clips and 1M video frames at 240 fps with a resolution of $1280 \times 720$, aiming to foster research in human-centric VFI.

\subsection{Video frame interpolation methods \label{2.2}}  
Existing VFI methods could be generally classified into flow-agnostic and flow-based methods.

\textbf{Flow-agnostic approaches} model VFI without explicit intermediate motion  representation. \emph{Phase-based} methods \cite{meyer2015phase, meyer2018phasenet} directly predict the phase decomposition of the intermediate frame, but can only handle motion within a limited range. \emph{Kernel-based} methods are the mainstream approach in this category, which typically aims to estimate intermediate frames by learning adaptive kernels to convolve input frames \cite{niklaus2017video,niklaus2017video2}. Over the years, numerous improvements are proposed in this field, including using deformable convolution \cite{cheng2020video,cheng2021multiple}, formulating interpolated motion estimation as classification \cite{peleg2019net}, blending deep features \cite{gui2020featureflow}, introducing dual-frame adversarial loss \cite{lee2020adacof}, performing channel attention \cite{choi2020channel} and utilizing 3D space-time convolutions \cite{kalluri2023flavr}. Recently, Shi \textit{et al.} \cite{shi2022video} introduced a Transformer-based framework to model long-range dependencies with the aid of attention mechanisms. By directly hallucinating pixel values, these methods tend to generate blurry results and artifacts, especially in fast-moving scenes \cite{lu2022video}.

\begin{figure*}[ht]
    \centering
    \includegraphics[width=\textwidth]{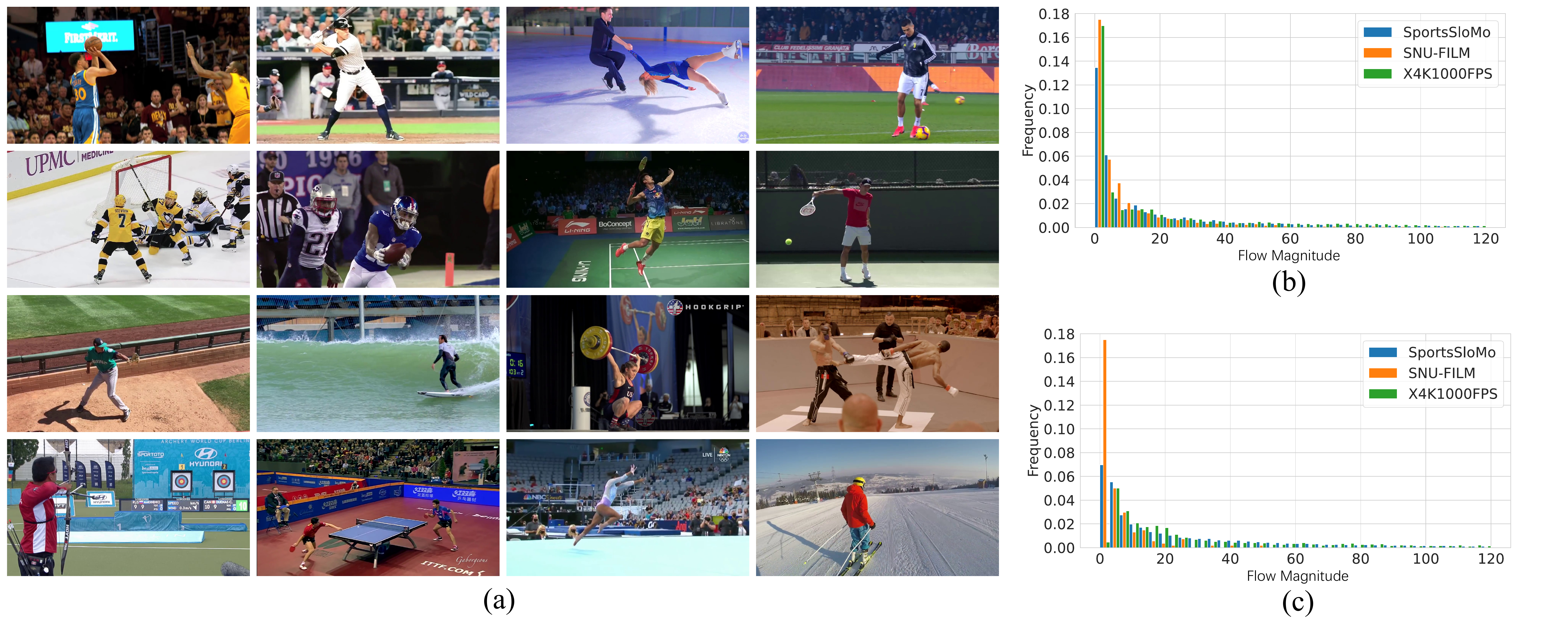}
    \caption{\textbf{The SportsSloMo dataset.} (a) Sampled frames, covering various sports categories and challenging human-centric content for VFI; (b) Histogram of flow magnitude of all pixels in the dataset; (c) Histogram of mean flow magnitude of all images in the dataset.}
    \vspace{-10pt}
    \label{fig:dataset}
\end{figure*}

\begin{figure}[htbp]
\centering
    \includegraphics[width=0.5\linewidth]{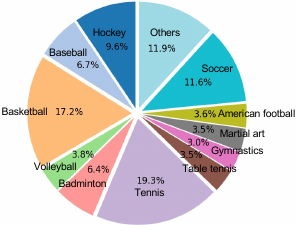}
\caption{\textbf{Distribution of different sports categories} in our SportsSloMo benchmark.}
\vspace{-10pt}
\label{fig:pie}
\end{figure}

\textbf{Flow-based approaches} currently serve as a promising direction in VFI. Generally speaking, flow-based method takes a paradigm of a two-stages pipeline: (1) flow estimation, and (2) frame synthesis. They first estimate optical flow between input frames, and then synthesize intermediate frames using image warping \cite{jaderberg2015spatial}. As a representative work, SuperSlomo \cite{jiang2018superslomo} by Jiang \textit{et al.} adopted a skip-connected U-Net to estimate bi-directional optical flows under the assumption of linear motion. Quadratic \cite{xu2019quadratic} and cubic \cite{chi2020all,tulyakov2022time} trajectory assumptions have also been made to approximate intermediate motion. Recent work has explored various techniques to improve intermediate flow estimation and interpolation accuracy, including forward-warping via softmax splatting \cite{niklaus2020softmax,hu2022manytomany}, voxel flow \cite{liu2017video}, cycle consistency loss\cite{reda2019unsupervised, liu2019deep}, task-oriented flow distillation loss \cite{kong2022ifrnet}, Gram matrix loss \cite{reda2022film}, implicit neural function \cite{chen2022videoinr}, occlusion mask \cite{bao2019memc}, anchor points alignment \cite{sim2021xvfi}, privileged distillation \cite{huang2022real}, and pyramid recurrent flow estimation \cite{jin2023unified}. Considering additional information like contextual maps \cite{niklaus2018context}, depth maps \cite{bao2019depth} and auxiliary visual information from event cameras \cite{tulyakov2021time,he2022timereplayer,chen2023revisiting} can also further improve interpolation accuracy. Park \textit{et al.} employed symmetric bilateral motion field estimation, and further improved intermediate motion estimation accuracy through asymmetric bilateral motion field \cite{park2021asymmetric}. Lu \textit{et al.} \cite{lu2022video} leveraged Transformer architecture \cite{vaswani2017attention} to model long-term dependency. Jin \textit{et al.} \cite{jin2023enhanced} proposed a novel bi-directional motion estimator in a pyramid structure. Zhang \textit{et al.} \cite{zhang2023extracting} proposed a novel feature extraction strategy to combine motion and appearance information via a hybrid CNN and Transformer architecture.

In a nutshell, although previous methods have proposed successful designs to handle complex motion and occlusion, none of them is carefully designed for human-centric scenes. As discussed in Sec. \ref{1}, human-centric VFI is confronted with various challenges including dynamic pose variation, complex human motion patterns and occlusion in crowded scenes. Additionally, accurately synthesizing fine details such as facial expressions and hand gestures can be challenging. To this end, we propose to consider human-aware loss terms through incorporating extra supervision to both human keypoints detection and segmentation in the panoptic setting. 

\section{SportsSloMo Benchmark \label{3}}
A myriad of benchmark datasets for video frame interpolation are available, including Middlebury \cite{baker2011database}, GoPro \cite{nah2017deep}, UCF101 \cite{soomro2012ucf101}, DAVIS \cite{perazzi2016benchmark}, Adobe240fps \cite{su2017deep}, Vimeo90K \cite{xue2019video}, SNU-FILM \cite{choi2020channel} and X4K1000FPS \cite{sim2021xvfi}. 
But none of these them focuses on human-centric VFI, \eg, in sports scenes. 
This limits the study of VFI methods targeted for human-centric applications such as enhanced entertainment experiences, commercial deployment, etc. 

\begin{figure*}[t]
\begin{center}
    \includegraphics[width=0.68\linewidth]{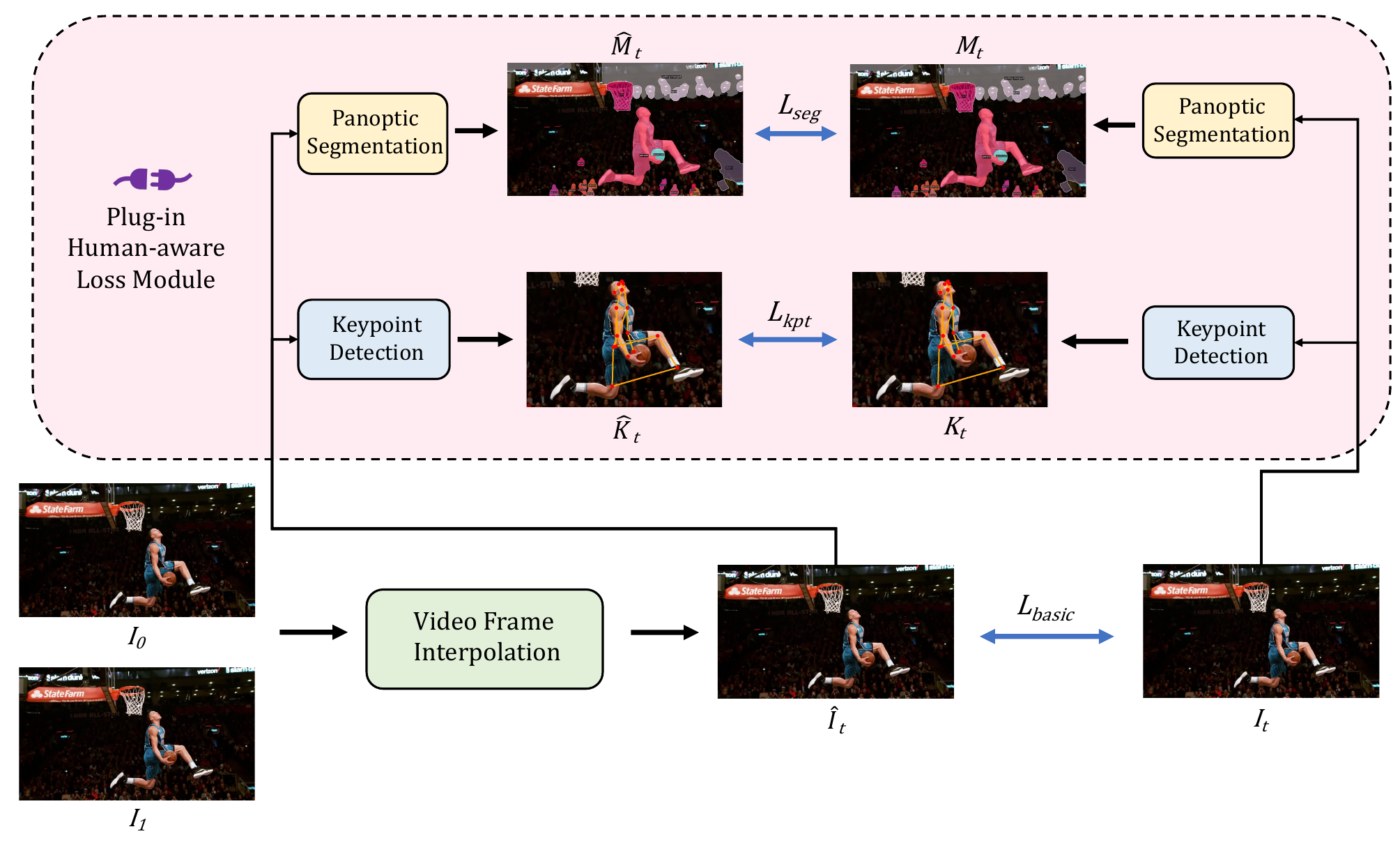}
\end{center}
\vspace{-16pt}
\caption{\textbf{Illustration of our proposed human-aware loss terms,} consisting of losses about human segmentation in the panoptic setting and keypoint detection.}
\vspace{-8pt}
\label{fig:pipeline}
\end{figure*}

To bridge this gap and to foster future research, we propose \textbf{SportsSloMo}, a challenging dataset consisting of high-resolution ($\geq$ 720p) sports videos crawled from YouTube under the Common Creative License. 
Careful curation has been taken to remove the unwanted content in the videos, including advertisement, transition frames, changes of shot, flashing lights, and non-slow-motion content.
Specifically, we first conduct human detection on all the videos by utilizing Yolov3 \cite{redmon2018yolov3} and remove video frames without detected humans. 
Second, we remove frames with flashing lights by setting a threshold about the brightness change between consecutive frames (large brightness change indicates the existence of a flashing light).  
Third, we use RAFT \cite{teed2020raft} to measure motion magnitude and set a threshold to discard non-slow-motion video segments. 
Finally, we carefully curate the the extracted clips manually to keep only high-quality self-consistent videos.
With such semi-automatic curation, we end up having a set of short slow-motion clips of 9 frames, where each frame has a spatial resolution of at least 1280 $\times$ 720.
The first and last frames are used as input for a VFI method and the 7 intermediate frames as ground truths for model training and evaluation, approximately corresponding to converting a video of 30-fps (frames per second) to 240-fps.

In total, we collect 131,464 video clips and 1.2M individual video frames from 259 raw YouTube videos, which covers 22 various sports categories in our dataset with different content and motion patterns, including hockey, baseball, skating, basketball, running, volleyball, etc.
Fig.~\ref{fig:pie} shows the proportion of different sports categories. 
Visualizations of randomly sampled video frames are available in Fig.~\ref{fig:dataset}.
We refer the readers to the supplementary material for more results.

As shown in Table~\ref{tab:dataset_statistics}, compared with existing datasets, not only our dataset is tailored for human-centric scenarios, but also surpasses others in terms of scale, resolution, and frame rate.
Fig. \ref{fig:dataset} demonstrates the histogram of flow magnitude of all pixels and mean flow magnitude of all images in our dataset (calculated using GMFlow \cite{xu2022gmflow}), with a comparison with the widely-used SNU-FILM~\cite{choi2020channel} dataset (which has the same spatial resolution as ours) and the recently introduced X4k1000FPS \cite{sim2021xvfi} dataset. 
As can be seen, our SportsSloMo benchmark contains more large-displacement motion compared to the SNU-FILM dataset. For instance, for the avarege flow magnitude on the image-level (Fig.~\ref{fig:dataset}(c)), our dataset has way more images, whose mean flow magnitude is greater than 20 pixels. And we can also observe that both our SportsSloMo and X4K1000FPS datasets contain large motions.

We split our proposed SportsSloMo dataset into \emph{train} and \emph{test}, containing 115,421 and 16,043 video clips, respectively. 
For each sports category, videos are split into \emph{train} and \emph{test} \emph{without intersection}, so that the test videos are completely unseen during training. 

\section{Human-aware Video Frame Interpolation}

\subsection{Overview} 
\label{subsec:overview_pipeline}

Given two input frames $I_0$ and $I_1$, the goal of VFI is to synthesize the intermediate frame $I_t$ at an intermediate time step $t \in (0,1)$. 
A VFI method needs to find the pixel-wise correspondences between $I_0$ and $I_1$ and adaptively fuse corresponding pixels to synthesize each pixel in $I_t$.
Flow-based approaches, such as SuperSloMo~\cite{jiang2018superslomo} and EBME~\cite{jin2023enhanced}, usually explicitly model the correspondences as optical flow, whereas flow-agnostic methods, such as AdaCof~\cite{lee2020adacof}, use learned kernels to process the visual correspondence implicitly.

To supervise the network training, the reconstruction error of the intermediate frame is used as the loss.
For instance, in~\cite{jin2023enhanced}, the weighted sum of the Charbonnier loss \cite{charbonnier1994two} $L_{char}$ and census loss \cite{meister2018unflow} $L_{cen}$ between the ground truth intermediate frame $I_t$ and synthesized frame $\hat{I}_t$ are adopted
\begin{equation}
    \begin{aligned}
        L_{basic} = L_{char}(I_t - \hat{I}_t) + \lambda_{cen} \cdot L_{cen}(I_t, \hat{I}_t),
    \end{aligned}
\end{equation}
where $L_{char}(x) = (x^2 + \epsilon^2)^\alpha, \alpha = 0.5, \epsilon = 10^{-6}$. 

Such reconstruction loss, however, may not provide enough supervision for human-centric VFI due to the challenges of highly deformable human motion and frequent occlusion. 
To address these challenges, we propose to incorporate $L_{seg}$ and $L_{kpt}$ into the network supervision based on human segmentation in the panoptic setting and keypoint detection, respectively, enforcing the model to produce high-quality synthesis results over human boundaries and along the keypoint trajectories. 
The final loss is
\begin{align}
L=L_{basic} + \lambda_{seg}L_{seg} + \lambda_{kpt}L_{kpt}.
\end{align}
It is worth noting that our proposed human-aware loss terms are flexible, which can be plugged into other flow-based (\eg, SuperSloMo~\cite{jiang2018superslomo} and EBME~\cite{jin2023enhanced}) or flow-agnostic VFI models (\eg, AdaCoF~\cite{lee2020adacof}), as shown in Section~\ref{sec:exp}.

\subsection{Human Segmentation Loss} 
In human-cenrtic scenarios, 
accurate estimation of human body boundaries is crucial.
In regions where the body movement is complex or there are heavy occlusions,  
inaccuracies in the estimated body boundaries may lead to visible artifacts in the synthesized video frame. 
To this end, we propose to incorporate human segmentation as extra supervision to improve the synthesis of intermediate video frames.
Human segmentation masks directly tell where human body boundaries are, as shown in Fig.~\ref{fig:kptseg}, helping to reduce ghost effects. 
Though instance and panoptic segmentation could both serve for this purpose, we empirically find that the panoptic segmentation models tend to yield better interpolation results than instance segmentation counterparts.

Specifically, we adopt state-of-the-art Mask2Former \cite{cheng2022masked} trained on COCO panoptic dataset \cite{kirillov2019panoptic} with the Swin-L backbone \cite{liu2021swin} from Detectron2 \cite{wu2019detectron2} to generate panoptic segmentation masks $M_t$ from the ground-truth intermediate frame $I_t$. 
During training, the panoptic segmentation results $\hat{M}_t$ of the synthesized intermediate frame $\hat{I_t}$ obtained using the same Mask2Former model is compared with $M_t$.
Ideally, if the synthesized intermediate frame $\hat{I}_t$ is identical to the ground truth $I_t$, $\hat{M}_t$ should be the same as $M_t$.
By comparing their difference, the human segmentation loss enhances the consistency of the human body boundaries between $I_t$ and $\hat{I_t}$, which helps reduce the ghost effects. 
We follow the loss function of Mask2Former and use the binary cross-entropy loss $L_{ce}$ and the dice loss $L_{dice}$ \cite{milletari2016v} for the human segmentation loss
\begin{equation}
    \begin{aligned}
        L_{seg} = \lambda_{ce} L_{ce}(\hat{M}_t, M_t) + \lambda_{dice} L_{dice}(\hat{M}_t, M_t),
    \end{aligned}
    \label{eq:aux_seg_loss}
\end{equation}
where we set $\lambda_{ce} = 5.0$ and $\lambda_{dice} = 5.0$.
The panoptic segmentation model is frozen during training and the gradient of this loss will be backpropagated to a VFI model only in a way similar to the perceptual loss~\cite{johnson2016perceptual}, enforcing the model to produce high-quality synthesis results.

\begin{figure}[t]
\centering
    \includegraphics[width=0.9\linewidth]{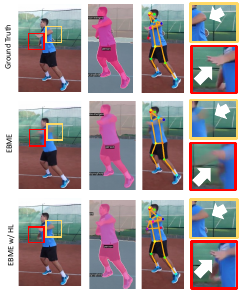}
\caption{\textbf{VFI results with human-aware loss (HL) terms.}
We can see better human segmentation, keypoint detection, and interpolation result can be obtained enhanced by the human-aware loss.
}
\vspace{-10pt}
\label{fig:addi}
\end{figure}

\begin{figure*}[htbp]
\centering
    \includegraphics[width=0.9\linewidth]{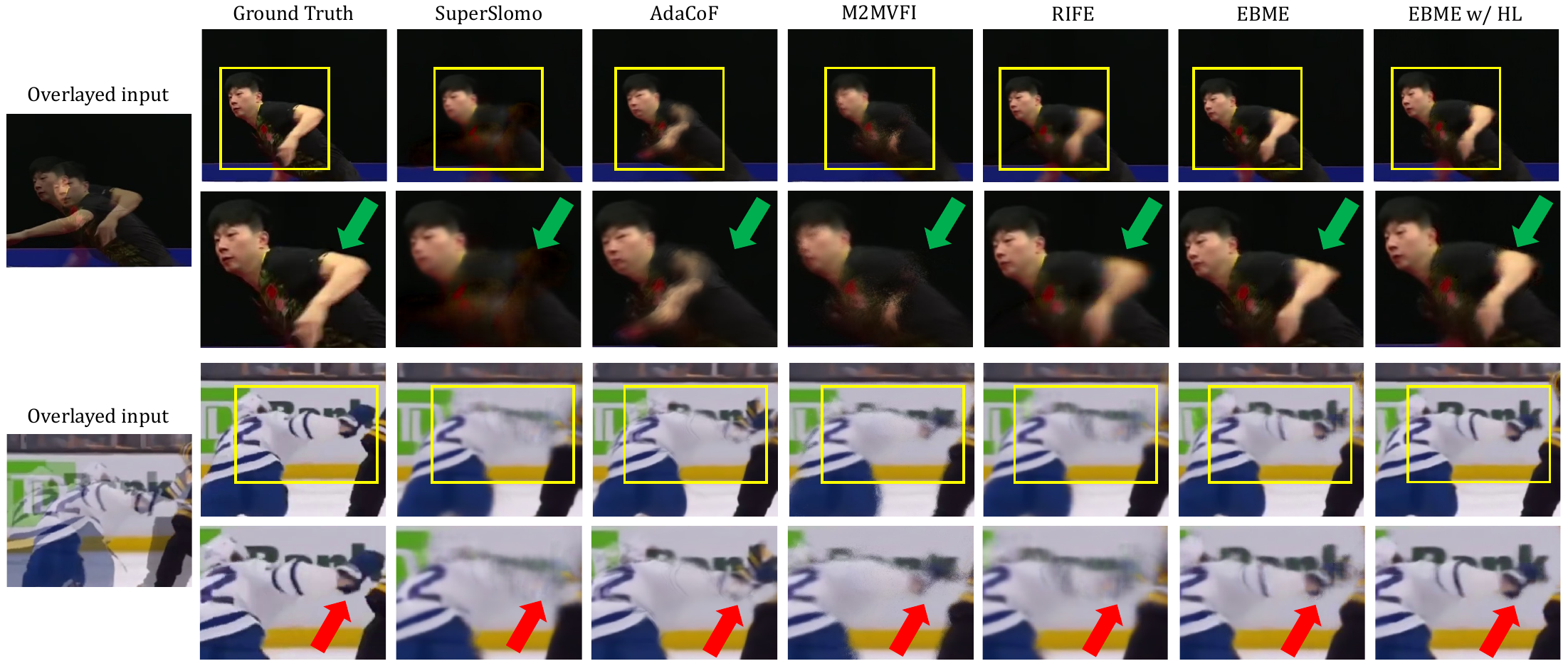}
\caption{\textbf{Qualitative comparisons on SportsSloMo dataset.} With the enhancement of the human-aware (HL) losses, EBME achieves better interpolation results. 
}
\vspace{-10pt}
\label{fig:finalquali}
\end{figure*}

\subsection{Human Keypoint Detection Loss} 
In addition to segmentation, we also consider the human keypoints estimation as an extra supervision. 
Accurate estimation of human keypoints is also crucial for human-centric VFI, which are indicative about the position of each body part, providing additional cues for motion estimation. 
Hence, by incorporating the supervision of human keypoints, we can better preserve human motions by enforcing coherent motion trajectories and guide the interpolation process to generate more plausible intermediate frames. 

Specifically, we adopt the state-of-the-art human pose estimation model ViTPose \cite{xu2022vitpose} trained on COCO \cite{lin2014microsoft}, with the ViT-L backbone \cite{dosovitskiy2020image} initialized with MAE \cite{he2022masked} pre-trained weights. We first use an off-the-shelf human detector YOLOv8 \cite{Jocher_YOLO_by_Ultralytics_2023} to detect person instances in the ground-truth intermediate frame $I_t$.
We then employ ViTPose to estimate the keypoints heatmap $K_t$ containing the locations of each joint.
Similar to the human segmentaiton loss, we estimate the keypoints heatmap $\hat{K}_t$ using the same ViTPose model on the synthesized intermediate frame $\hat{I}_t$.
The differences between $K_t$ and $\hat{K}_t$ are then used as the supervision to improve VFI results.
We follow the loss function of ViTPose and use the MSE (mean square error) loss over heatmaps as the human keypoint loss $L_{kpt}$. 
\begin{align}
    L_{kpt} = MSE(K_t, \hat{K}_t).
\end{align}
The ViTPose model is frozen during training and the gradient of this loss will be backpropagated to a VFI model to encourage high-quality intermediate frame synthesis so that the estimated human keypoints are close to the ones obtained from the true intermediate frame.

\section{Experiments}
\label{sec:exp}
\subsection{Setup}
\noindent \textbf{VFI models.} We benchmark existing VFI approaches by re-training several state-of-the-art methods on our SportsSloMo dataset using their publicly available code, including SuperSlomo \cite{jiang2018superslomo}, AdaCoF \cite{lee2020adacof}, M2MVFI \cite{hu2022manytomany}, RIFE \cite{huang2022real}, EBME \cite{jin2023enhanced}, UPR-Net \cite{jin2023unified}, and EMA-VFI \cite{zhang2023extracting}. 
AdaCoF is a flow-agnostic VFI model and the rest are flow-based. 
All these models are trained from scratch on our SportsSloMo dataset. 

\begin{table}[t]
    \centering
    \caption{\textbf{Quantitative results of different approaches on the SportsSloMo dataset.}
    As can be seen, our proposed human-aware loss (HL) terms consistently improve all VFI models' performance. 
    }
    \label{tab:sports_quantitative}
    \renewcommand{\tabcolsep}{4pt}
    \small
    \begin{tabular}{c|c|c|c|c}
    \toprule
     Method & Venue & PSNR$\uparrow$ & SSIM$\uparrow$ & IE$\downarrow$ \\
    \midrule
     SuperSlomo \cite{jiang2018superslomo} & CVPR 2018 & 29.77 & 0.910 & 9.14\\
     SuperSloMo + HL & - & \textbf{30.24} & \textbf{0.917} & \textbf{8.46} \\
    \midrule
      AdaCoF \cite{lee2020adacof} & CVPR 2020  & 28.79 & 0.926 & 5.84\\
      AdaCoF + HL & -  & \textbf{28.94} & 0.926 & \textbf{5.72}\\
    \midrule
     M2MVFI \cite{hu2022manytomany} & CVPR 2022 & 29.03 & 0.935 & 5.16\\
     M2MVFI + HL & - & \textbf{29.29} & \textbf{0.936} & \textbf{5.07} \\
    \midrule
     RIFE \cite{huang2022real} & ECCV 2022 & 29.69 & 0.931 & 5.25\\
     RIFE + HL & - & \textbf{29.87} & \textbf{0.933} & \textbf{5.16} \\
    \midrule
     EBME \cite{jin2023enhanced} & WACV 2023 & 30.15 & 0.941 & 4.66\\
     EBME + HL & - & \textbf{30.48} & \textbf{0.944} & \textbf{4.40}\\
    \midrule
     UPR-Net \cite{jin2023unified} & CVPR 2023 & 30.25 & 0.945 & 4.56\\
     UPR-Net + HL & - & \textbf{30.50} & 0.945 & \textbf{4.38} \\
     \midrule
     EMA-VFI \cite{zhang2023extracting} & CVPR 2023 & 30.70 & 0.949 & 4.28\\
     EMA-VFI + HL & - & \textbf{30.75} & \textbf{0.952} & \textbf{4.19} \\
    \bottomrule
    \end{tabular}
\end{table}

We also incorporate our proposed human-aware loss terms into these models to validate the effectiveness of explicitly taking human priors into account to improve human-centric VFI.

\noindent \textbf{Evaluation metrics.} Following previous VFI methods \cite{jiang2018superslomo,huang2020rife}, we adopt the widely-used signal-to-noise ratio (PSNR), structural similarity (SSIM) \cite{wang2004image}, and interpolation error (IE) \cite{baker2011database} to evaluate the interpolation results. For PSNR and SSIM, higher indicates better performance. And for IE, the lower the better.

\noindent \textbf{Implementation details.} 
For the human segmentation and keypoint loss terms, to avoid the influence of Batch Normalization (BN) layers on the training phase, we replaced the BN layers with frozen Batch Normalization (frozen BN) layers.
We use the default hyperparameters provided by each VFI model.
Training and evaluation are both conducted on 8 NVIDIA RTX A6000 GPUs. 

During training, we randomly sample a frame between the first and the ninth (last) frame for VFI methods supporting arbitrary time frame interpolation like SuperSlomo \cite{jiang2018superslomo}, M2MVFI \cite{hu2022manytomany}, RIFE \cite{huang2022real}, EBME \cite{jin2023enhanced}, UPR-Net \cite{jin2023unified}, and EMA-VFI \cite{zhang2023extracting}. 
For VFI methods that can only synthesize the middle frame like AdaCoF \cite{lee2020adacof}, we randomly generate triplets within the nine frames such that the target frame is always in the middle of the input two frames. 
During evaluation, arbitrary-time-supported VFI methods take the time step as input and interpolate every intermediate frame, while single-frame methods recursively generate each intermediate frame. 

\subsection{Benchmarking Existing Methods}

Table \ref{tab:sports_quantitative} shows quantitative comparisons between existing VFI methods on our proposed SportsSloMo dataset. 
We can see that EMA-VFI \cite{zhang2023extracting} performs the best interms of all three evaluation metrics.
It is worth noting that the performance for each model decreases compared with results on popular datasets. 
For instance, two top-performing approaches on SportsSloMo, EBME~\cite{jin2023enhanced} and EMA-VFI~\cite{zhang2023extracting} produce PSNR scores of 30.15 and 30.70, respectively, compared with 36.19 and 36.64 on the Viemo90K~\cite{xue2019video} as well as 30.64 and 30.94 on the hard split of the SNU-FILM~\cite{choi2020channel} benchmarks.
It highlights the difficulty of our benchmark and suggests significant challenges need to be addressed.
We provide more details in the supplementary material.

 \subsection{Enhanced Human-aware VFI}
 Table \ref{tab:sports_quantitative} also demonstrates the effectiveness of our proposed human-aware losses. 
 Specifically, we incorporate the proposed human segmentation and keypoint losses into each of the VFI models.
 As can be seen, they consistently improve the performance of every single VFI method for all three evaluation metrics.
 Specifically, in terms of PSNR, our proposed human-aware losses lead to improvement of 1.6\% and 1.1\% for SuperSloMo~\cite{jiang2018superslomo} and EBME~\cite{jin2023enhanced}, respectively.
 In terms of IE, the improvement for these two methods are 7.4\% and 5.6\%, respectively.
 Qualitatively, we can see from Fig.~\ref{fig:addi} and Fig.~\ref{fig:finalquali}, our proposed human-aware loss terms can improve interpolation results for the highly deformable arms of the athletes and the background under occlusions.
 
\subsection{Ablation Study}
In this section, we present ablation study on SuperSloMo~\cite{jiang2018superslomo} and EBME \cite{jin2023enhanced} to analyze the design choices of our proposed human-aware losses in Table~\ref{tab:ablation}.
As can be seen, both human segmentation and keypoint detection losses can successfully improve the performance of SuperSloMo and EBME. 
This justifies our motivation to leverage human-aware priors to improve the human-centric VFI. 
By combining these two losses, the largest performance gain can be obtained for both methods.
Furthermore, from Fig.~\ref{fig:addi}, we can clearly see that better interpolation results can be obtained around the elbow and hand. 

\begin{table}[t]
    \centering
    \caption{\textbf{Effectiveness of different auxiliary loss terms.}} 
    \label{tab:ablation}
    \renewcommand{\tabcolsep}{4pt}
    \small
    \begin{tabular}{c|c|c|c|c|c}
    \toprule
         \multirow{2}{*}{Model} & segmentation & keypoint & \multirow{2}{*}{PSNR$\uparrow$} & \multirow{2}{*}{SSIM$\uparrow$} & \multirow{2}{*}{IE$\downarrow$} \\
     & loss & loss & & & \\
     \midrule
     \multirow{4}{*}{SuperSloMo} & \xmark & \xmark & 29.77 & 0.910 & 9.14\\
      & \cmark & \xmark & 30.19 & 0.916 & 8.56\\ 
     & \xmark & \cmark & 30.22 & 0.916 & 8.49\\
     & \cmark & \cmark & \textbf{30.24} & \textbf{0.917} & \textbf{8.46} \\
     \midrule
     \multirow{4}{*}{EBME} & \xmark & \xmark & 30.15 & 0.941 & 4.66\\
      & \cmark & \xmark & 30.26 & 0.942 & 4.61\\
     & \xmark & \cmark & 30.34 & 0.943 & 4.46\\
     & \cmark & \cmark & \textbf{30.48} & \textbf{0.944} & \textbf{4.40}\\
    \bottomrule
    \end{tabular}
\end{table}

\subsection{Limitations and Discussions}
\label{subsec:flow_loss}
While we have introduced human-aware loss terms to handle challenging human-centric scenarios in our proposed dataset, there still remains noteworthy limitations for future exploration.

First of all, for human-centric scenes, the presence of large, complex, and highly deformable human motions and occlusion by crowds make finding correspondences of humans (\ie, optical flow) between input frames a challenging task. 
Recent studies \cite{huang2022real, kong2022ifrnet} have demonstrated the effectiveness of knowledge distillation \cite{hinton2015distilling} to improve the intermediate optical flow estimation. 

We also explored this loss to incorporate extra supervision for flow-based VFI methods (\eg, SuperSloMo~\cite{jiang2018superslomo}, EBME~\cite{jin2023enhanced}).
Specifically, we use the optical flow results obtained from state-of-the-art model GMFlow \cite{xu2022gmflow}, which is pre-trained on the optical flow benchmarks~\cite{Dosovitskiy:2015Flownet,mayer2016a,Butler:ECCV:2012}, as the teacher network to perform flow distillation. 
We found, however, mixed results on our SportsSloMo benchmark.
While it indeed improves the performance for SuperSlomo~\cite{jiang2018superslomo} (PSNR: 29.77 \emph{vs.} 30.18), it hurts the performance for EBME~\cite{jin2023enhanced} (PSNR: 30.15 \emph{vs.} 29.85). 
We also fine-tune the GMFlow model on the human flow dataset~\cite{ranjan2020learning}, consisting of synthetic human figures overlaied on top of real world images as background.
However, we observed significant domain gap between the synthetic training images and real-world test images, which lead to poor optical flow estimation results. 
How to improve human-centric optical flow to improve VFI is still a challenging problem.

Second, we only utilize 2D cues of human bodies in our proposed loss terms to enhance VFI.
Recently, there has been significant advance made in 3D human body reconstruction from videos, for instance, \cite{goel2023humans,ye2023slahmr,rajasegaran2023benefits,Kocabas_PARE_2021}, considering both the spatial and temporal information.
By lifting 2D videos into the 3D space, occlusions may be better solved. 
More advanced loss terms involving 3D human reconstruction is worth exploring. 
We leave this as future work to explore.

Finally, in the SportsSloMo dataset, various sports involve fast-moving balls and sports equipment. 
In this paper, we only consider human boundary and body parts for motion estimation. 
Fast-moving objects, however, in these scenarios are also challenging for VFI. 
More effort is needed for such a problem.

\section{Conclusion}
In this paper, we introduced a new benchmark, \textbf{SportsSloMo}, focusing on human-centric video frame interpolation. 
Our benchmark contains 130K video clips and more than 1M video frames obtained from high-resolution ($\geq$ 720p) slow-motion sports videos crawled from YouTube with careful curation.
Due to the complex, highly deformation human motion and frequent occlusion, this benchmark imposes significant challenges to existing VFI models.
To enhance their accuracy, we introduce human-aware loss terms to improve existing methods, where the supervision of human segmentation and keypoints detection are incorporated.
Our loss terms are model agnostic and have been successfully applied to seven existing VFI methods, leading to better accuracy consistently.
Our benchmark dataset and code will be publicly released to foster future research in the new exciting direction of human-centric VFI.

\section*{Acknowledgement} We thank Thuy-Tien Bui and Devroop Kar for their help on data collection. This work was partially supported by the National Science Foundation under Award IIS-2310254.

{\small
\bibliographystyle{ieee_fullname}
\bibliography{egbib}
}

\end{document}